\documentclass[conference]{IEEEtran}
\usepackage{amsmath,amssymb,amsfonts}
\usepackage{algorithmic}
\usepackage{graphicx}
\usepackage{textcomp}
\usepackage{xcolor}
\usepackage{microtype}
\usepackage{hyperref}
\usepackage{wrapfig}
\usepackage{todonotes}

\newcommand{\R}{\mathbb R} 
\newcommand{\F}{\mathcal F} 
\DeclareMathOperator{\Cov}{Cov}
\DeclareMathOperator{\Corr}{Corr}
\DeclareMathOperator{\argmin}{argmin}
\DeclareMathOperator{\rank}{rank}
\newcommand{\T}{\mathsf{T}}

\begin{document}
	
	\title{Characterization of Hemodynamic Signal by Learning Multi-View Relationships} 
	
	\author{\IEEEauthorblockN{1\textsuperscript{st} Authors}
		\IEEEauthorblockA{\textit{Department} \\
			\textit{name of organization (of Aff.)}\\
			City, Country \\
			email address}
	}
	\author{\IEEEauthorblockN{Eric Lei}
		\IEEEauthorblockA{\textit{Auton Lab} \\
			\textit{Carnegie Mellon University}\\
			Pittsburgh, USA \\
			elei@cs.cmu.edu}
		\and
		\IEEEauthorblockN{Kyle Miller}
		\IEEEauthorblockA{\textit{Auton Lab} \\
			\textit{Carnegie Mellon University}\\
			Pittsburgh, USA \\
			mille856@andrew.cmu.edu}
		\and
		\IEEEauthorblockN{Michael R.\ Pinsky}
		\IEEEauthorblockA{\textit{School of Medicine} \\
			\textit{University of Pittsburgh}\\
			Pittsburgh, USA \\
			pinsky@pitt.edu}
		\and
		\IEEEauthorblockN{Artur Dubrawski}
		\IEEEauthorblockA{\textit{Auton Lab} \\
			\textit{Carnegie Mellon University}\\
			Pittsburgh, USA \\
			awd@cs.cmu.edu}
	}
	
	\maketitle
	\thispagestyle{plain}
	\pagestyle{plain}
	
	\begin{abstract}
		\textit{Objective:} Multi-view data are increasingly prevalent in practice. It is often relevant to analyze the relationships between pairs of views by multi-view component analysis techniques such as Canonical Correlation Analysis (CCA). However, data may easily exhibit nonlinear relations, which CCA cannot reveal. We aim to investigate the usefulness of nonlinear multi-view relations to characterize multi-view data in an explainable manner. \textit{Methods:} To address this challenge, we propose a method to characterize globally nonlinear multi-view relationships as a mixture of linear relationships. A clustering method, it identifies partitions of observations that exhibit the same relationships and learns those relationships simultaneously. It defines cluster variables by multi-view rather than spatial relationships, unlike almost all other clustering methods. Furthermore, we introduce a supervised classification method that builds on our clustering method by employing multi-view relationships as discriminative factors. The value of these methods resides in their capability to find useful structure in the data that single-view or current multi-view methods may struggle to find. \textit{Results:} We demonstrate the potential utility of the proposed approach using an application in clinical informatics to detect and characterize slow bleeding in patients whose central venous pressure (CVP) is monitored at the bedside. Presently, CVP is considered an insensitive measure of a subject's intravascular volume status or its change. However, we reason that features of CVP during inspiration and expiration should be informative in early identification of emerging changes of patient status. We empirically show how the proposed method can help discover and analyze multiple-to-multiple correlations, which could be nonlinear or vary throughout the population, by finding explainable structure of operational interest to practitioners.
	\end{abstract}
	
	
	\section{Introduction}
	
	Increasingly prevalent in modern statistics and machine learning, multi-view data are relational datasets whose features can be partitioned into sets, called views. This structure enables exploitation of relationships between views to find insights that classical single-view methods cannot. Multi-view data can be found in many domains, including physiologic and clinical.
	Examples include genes and diseases~\cite{seoane2014canonical}, visuals and text~\cite{rasiwasia2010new}, emotions and personality disorders~\cite{sherry2005conducting}, and physiologic vital sign waveforms. 
	A common multi-view approach is to analyze correlation between views, often using Canonical Correlation Analysis (CCA) from classical statistics, a standard way to characterize multi-view relationships~\cite{hotelling1936relations}.
	CCA finds linear projections from each view into a shared latent space such that the projections have maximal correlation.
	According to~\cite{Bach2006}, the canonical, or latent, variables can be considered the basis of a generative model for the observed views.
	These variables often have some practical meaning, such as a certain combination of genes that corresponds to a combination of biological phenotypes~\cite{Witten2009}.
	They can then be used to analyze complex datasets in an interpretable fashion.
	A strict limitation of CCA, however, is its linearity. In many real datasets, the correlations between views could have globally nonlinear structure. Therefore, although CCA might have some relevancy, it could fail to capture important information.
	
	This work investigates the usefulness of nonlinear multi-view relationships to characterize multi-view data in an explainable manner. To do so, we propose a nonlinear method based on CCA that approximates globally nonlinear structure as cluster-wise linear. The method may be considered a  clustering algorithm for a mixture of CCA models. Intuitively, different parts of the data may have distinct patterns of correlation because important latent variables might differ between subsets of observations.
	For instance, certain subpopulations might express a gene combination differently, or distinct subsets of subjects might have a different physiological response to medical trauma.
	To learn this structure, our method, called Canonical Least Squares (CLS) clustering, identifies clusters of observations, but these clusters fundamentally differ from classical clustering methods because they are based on multi-view relationships rather than spatial relationships.
	This approach can be considered a form of correlation clustering, a class of clustering methods that groups  observations based on their correlation patterns~\cite{Zimek2009}.
	Additionally, we introduce a supervised classification method that extends CLS clustering. This method hypothesizes that multi-view relationships can serve as discriminative factors between classes, taking a novel approach to explicitly model these relations.
	Practical benefits of these methods stem from their capability to find interpretable structure in the data to explain their predictions. 
	
\begin{figure}[ht!]
		\centering
		\includegraphics[width=.36\textwidth]{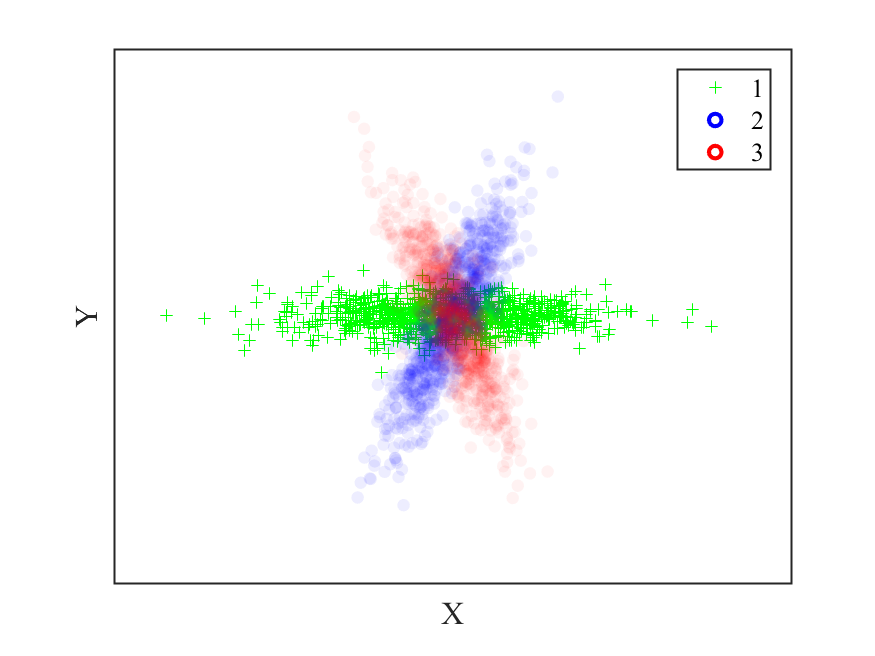}
		\caption{Ground truth clusters for an overlapping Gaussian dataset, drawn translucently to illustrate overlap.}
		\label{fig:star-true}
		\includegraphics[width=.36\textwidth]{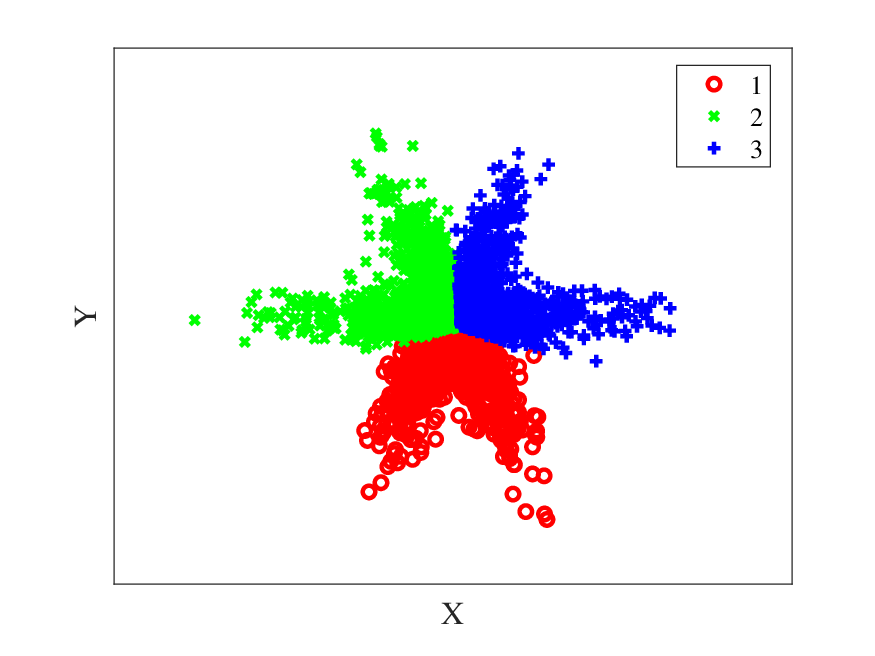}
		\caption{Clusters from $k$-means or spectral clustering for an overlapping Gaussian dataset.}
		\label{fig:star-km-sc}
		\includegraphics[width=.36\textwidth]{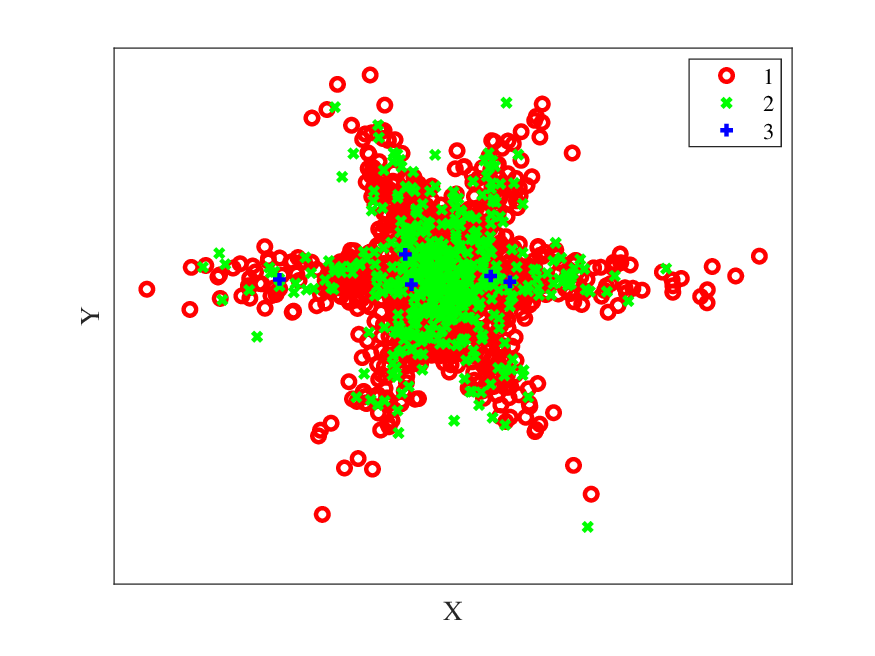}
	    \caption{Clusters from deep non-negative matrix factorization~\cite{Zhao2017} for an overlapping Gaussian dataset.}	
	    \label{fig:star-deepnmf}
		\includegraphics[width=.36\textwidth]{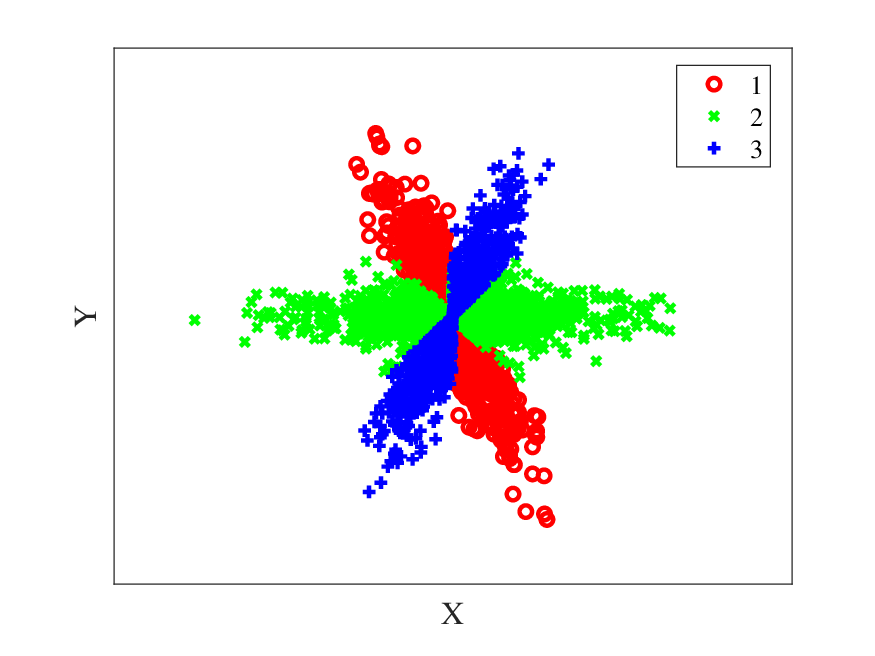}
    	\caption{CLS clusters for an overlapping Gaussian dataset.}
    	\label{fig:star-cls}
	\end{figure}

    The novelty of this approach is that it can easily find interesting multi-view structure with which single-view or current multi-view methods struggle. By ignoring relations between views or factoring them in only indirectly, the alternatives  risk overlooking key information. According to~\cite{Liu2013}, multi-view clustering strategies can usually be grouped into three categories. First, multiple views are integrated through the loss function, which includes the method in Fig.~\ref{fig:star-deepnmf}. Second, multi-view data are projected to a common subspace, in which any standard clustering algorithm is then applied. Third, a clustering solution is computed for each view individually, and then they are all fused to achieve a consensus.
    Every category, however, overlooks multi-view relationships or models them only implicitly.
    Even the second category of subspace learning, though most similar to the ideas we propose, ultimately operates on spatial relationships in some feature space by spatially clustering data points in the end.  In contrast, our methods define the inference task at hand explicitly through multi-view relationships by defining cluster variables as the relations themselves. In short, almost all current multi-view methods lack an additional layer of abstraction on top of views to truly separate themselves from single-view methods that project into a different feature space or apply a specific mode of regularization. We hypothesize that our work provides this layer.
	
	To illustrate our approach, consider the two-dimensional dataset in Fig.~\ref{fig:star-true}. Each axis corresponds to the projection of a view to one dimension. The data contain three spatially overlapping Gaussian clusters with different covariance structures. When $k$-means or Gaussian kernel spectral clustering is employed, the resulting clusters are displayed in Fig.~\ref{fig:star-km-sc}. As expected, they are contiguous in space, but they do not match the ground truth because of the overlap in  data. Next, the result of a popular multi-view clustering approach is shown in Fig.~\ref{fig:star-deepnmf}. Although this approach discovers overlapping clusters, it only finds two clusters, and they appear to be random samples from the original distribution. It finds clusters by applying a single-view approach to each view while enforcing agreement between the results. The issue with this procedure  is that all information about the clusters is lost in this dataset if considered from only one view.
	In contrast, Fig.~\ref{fig:star-cls} illustrates the clusters learned by CLS clustering. They closely resemble the ground truth despite the overlapping data, though the learned clusters do not overlap spatially.
	This example demonstrates that there are certain problems in which the data have an interesting structure that cannot be discovered by straightforward methods. This idea motivates our approach, 
	distinguished from more common methods by the ability to account for such more unusual structures that may be prevalent in practice.
	
	As an application, we consider an important area in medicine concerning the detection of internal bleeding in post-surgical patients, evaluating our clustering and classification methods on a dataset in this field. Such bleeding if undetected for prolonged intervals is associated with profound morbidity and even death. However, its overt signatures in gross-measured hemodynamic variables are often remarkably benign, until profound hypovolemia is present. Although invasive monitoring of central venous pressure (CVP) is common, raw CVP values or their changes are insensitive measures to assess a subject's intravascular volume status or its change~\cite{pinsky2005functional,kumar2004pulmonary,marik2013does}. However, CVP is also an essential component of the determinants of venous return to the heart defining cardiac output. Positive pressure breathing cyclically increases intrathoracic pressure as lung volume increases, causing CVP to also increase, transiently decreasing venous return. Thus, we reason that comparative aspects of the CVP pressure waveform signal between inspiration and expiration may be informative in assessing changes in intravascular volume status even if average CVP values are not. Since CVP is the most common invasive hemodynamic monitoring feature used clinically, and since occult hypovolemia due to   post-surgery hemorrhage is a common complication, we examined the ability of CLS to analyze CVP data to identify the onset and characterize the extent of bleeding in a clinically-relevant porcine model of fixed-rate blood loss.   
	
	Furthermore, in variety of clinical settings, it is common to encounter datasets of high complexity, reflecting the high variance in physiological responses.
	Although machine learning models have been applied successfully to a range of clinical scenarios, medical practitioners are wary of methods that lack explanatory power~\cite{Murdoch2013,krumholz2014,Holzinger2014,Obermeyer2016,Choi2016a}.
	Recently, a growing effort has been made to develop ``translucent box" methods that offer some extent of interpretability of the learned models and predictions they make, improving face validity of data-driven analytics in clinical applications.
	Our approach may be viewed as a contribution to this line of effort.
	
	The main contributions of this paper are:
	\vspace*{-4pt}
	\begin{enumerate}
		\itemsep-.2em 
		\item A clustering method for multi-view data based on relationships between views.
		\item A multi-view classification method that employs multi-view relationships as discriminative factors.
		\item A demonstration of the proposed approach in a practical context of clinical importance.
	\end{enumerate}
	
	The paper is organized as follows: Section~\ref{sec:rel} summarizes related work; Section~\ref{sec:cca} gives necessary background on CCA; Section~\ref{sec:cls} derives the CLS clustering and classification methods; Section~\ref{sec:results} describes experimental results; Section~\ref{sec:discussion} discusses interpretations of results and comparisons to other methods; and Section~\ref{sec:conclusion} concludes the paper.
	
	\section{Related Work}
	\label{sec:rel}
	
	The concept of multi-view correlation clustering has been explored in the past, albeit with limited scope.
	A simplified version of our work is cluster-wise linear regression~\cite{spath1982fast}, a  method for clustering the observations in a single-output regression dataset. Like $k$-means, this method is greedy and iterative and alternates between two steps. Given cluster labels, it fits a  linear regression to each cluster. Given regression coefficients, it assigns each observation to the cluster whose regression residual is the smallest for that observation. It is simple to show that this method is a special case of CLS clustering in which the regression inputs are one set of variables and the regression output by itself is the other set---i.e., one view is univariate.
	Another interesting approach is explored by~\cite{klami2008probabilistic} and~\cite{rey2012copula}. \cite{klami2008probabilistic} establish a probabilistic generative modeling framework to allow Bayesian inference. They do so by proposing a model of probabilistic families for finding dependency and give a general clustering algorithm for this family. CCA is shown to be a special case. A key assumption is that a linearly transformed Gaussian latent variable produces the variation in the data. However, there may be severe model mismatch when this assumption was violated. To remedy this behavior,~\cite{rey2012copula} deploy a copula mixture model to the framework, enabling them to model mixtures of CCA, similar to the clustering setup in this work. A Bayesian clustering algorithm is proposed and shown to perform well on synthetic and real datasets. However, a disadvantage of this approach is that it requires a prior distribution to be specified for every feature, which could require a vast number of hyperparameters and introduces the strong possibility of model mismatch.
	In addition, in~\cite{Blaschko2008}, a multi-view clustering method called Correlational Spectral Clustering is proposed based on kernel CCA (KCCA).
	It  runs KCCA on the kernel matrices of two views and then runs $k$-means on the latent variables in one view. The authors state that this method generalizes spectral clustering to arbitrary kernels and paired data. A notable distinction from our work is that KCCA cluster assignments for test observations depend on only one view. For example, if it were run on the data in Fig.~\ref{fig:star-true}, it would find clusters divided along vertical lines.  Hence, the clusters have limited dependence on correlation between views.
	
	Additionally, there has been substantial work on multi-view clustering that operates on spatial relationships.
	Many methods learn a subspace and apply a single-view clustering algorithm. For example, a method by \cite{Chaudhuri2009} uses CCA to find the subspace spanned by the means of mixture components.
	In \cite{Cao2015}, a subspace learning approach induces diversity between projections in each view.
	Other methods regularize solutions in each view to become similar during training. For instance,
	\cite{Kumar2011} propose a kind of multi-view spectral clustering, employing co-regularization to enforce agreement between clusterings in different views.
	A different spectral approach was presented by \cite{Wang2013}, applying structured sparsity to weight features in different views by their importance. 
	Some methods enforce similarity between views by preserving graph structure, such as \cite{Nie2017}.  
	Another popular direction in this category is clustering by nonnegative matrix factorization, a method introduced by \cite{Liu2013} that searches for factorizations that give compatible clusters across the views. This method has many extensions, such as \cite{Zong2017} and \cite{Zhao2017}.
	While interesting, all these approaches differ from ours because they ultimately cluster on spatial relationships while ours clusters explicitly on multi-view relationships.
	
	A related line of research is single-view correlation clustering. \cite{Zimek2009} considers the problem of clustering data based on patterns of correlation when the variables are not partitioned into two groups. Unlike CLS or CCA, however, this work assumes a single view; the correlation refers to correlation between all the variables, not just between two sets. The paper presents a diverse body of algorithms for the task.
	
	
	\section{Canonical Correlation Analysis}
	\label{sec:cca}
	In this section we summarize CCA, a useful starting point for understanding the proposed methods.
	A classical way to characterize relationships between views, CCA analyzes cross-covariance between two sets of variables that have aligned observations. 
	By performing CCA, one can understand how much variance in the sets can be explained by common factors.
	Let $X \in \R^{d_X}$ and $Y \in \R^{d_Y}$ be random vectors. 
	Without loss of generality, assume $\mathrm E[X]=\mathrm E[Y]=0$.
	Then CCA for the  $m$-th component solves the problem
	\begin{equation} \label{eqn:cca1}
	\begin{aligned}
	& \underset{u\in\R^{d_X},v\in\R^{d_Y}}{\max}
	& & \Corr(X^\T u, Y^\T v) \\
	& \text{\quad subject to}
	& & \Cov(Xu,Xu_i) = \Cov(Yv,Yv_i) = 0, \\
	& & & i=1,\dots,m-1.
	\end{aligned}
	\end{equation}
	Define $\Sigma_{XY}=\Cov(X,Y)$, $\Sigma_{XX}=\Cov(X)$, and $\Sigma_{YY}=\Cov(Y)$.
	This optimization is non-convex, but it  has a closed-form solution~\cite{hardoon2004canonical}: $u$ and $v$ are the respective $m$-th largest eigenvectors of
	\[A=\Sigma_{XX}^{-1} \Sigma_{XY} \Sigma_{YY}^{-1} \Sigma_{XY}^\T,\] 
	\[B=\Sigma_{YY}^{-1} \Sigma_{XY}^\T \Sigma_{XX}^{-1} \Sigma_{XY}.\]
	
	A standard reformulation~\cite{hardoon2004canonical} of the objective function in~(\ref{eqn:cca1}) relevant to our method is
	\begin{equation} \label{eqn:cca2}
	\underset{u\in\R^{d_X},v\in\R^{d_Y}}{\min} \mathrm{E} \left[\|Xu-Yv\|_2^2\right].
	\end{equation}

	\section{Canonical Least Squares Clustering}
	
	\label{sec:cls}
	In this section we develop our method to characterize nonlinear multi-view relationships called Canonical Least Squares (CLS) clustering. 
	The intuition is that globally nonlinear multi-view relationships can be approximated by cluster-wise linear relationships. Our method clusters observations such that each cluster corresponds to a separate linear relationship between views.
	We then describe how it can serve as the basis of supervised classification.
	First we explain CLS, an optimization problem analogous to CCA, to serve as the model of linear multi-view relationships. 
	Like CCA, it takes sets of variables $X$ and $Y$ and produces up to $m\le\min(d_X,d_Y,\rank(X^\T Y))$ pairs of vectors $(u,v)$ such that the components $X^\T u$ and $Y^\T v$ have some kind of relationship. Unlike CCA, this relationship is not of maximum correlation but of least squared error.
	Although a non-convex problem, CLS has a closed-form solution for the first component and a simple approximation to the other components.

	\subsection{First Components}
	First consider only the top pair of components ($m=1$).
	We redefine $X\in\R^{n\times d_X}$ and $Y\in\R^{n\times d_Y}$ as centered data matrices. Then~(\ref{eqn:cca2}) becomes

	\begin{equation*} 
	\begin{aligned}
	& \underset{u\in\R^{d_X},v\in\R^{d_Y}}{\min}
	& & \|Xu-Yv\|_2^2 \\
	& \text{\quad subject to}
	& & u^\T X^\T X u = v^\T Y^\T Y v =1.
	\end{aligned}
	\end{equation*}
	
	We propose the following modification, which has the same objective but different constraints:
	
	\begin{equation}
	\label{eqn:cls1} 
	\begin{aligned}
	& \underset{u\in\R^{d_X},v\in\R^{d_Y}}{\min}
	& & \|Xu-Yv\|_2^2 \\
	& \text{\quad subject to}
	& & v^\T v =1.
	\end{aligned}
	\end{equation}
	
	This optimization problem has a positive semidefinite objective but quadratic constraints, so it is not convex. We denote~(\ref{eqn:cls1}) CLS (for the first component).
	One major difference from CCA is the lack of $X$ or $Y$ in the constraints. 
	This difference enables CLS to form the building block of a clustering method with a well-defined optimization procedure, as will soon be explained.
	The other difference is the lack of $u$ in the constraints. 
	When only $v$ is constrained, the problem generalizes ordinary least squares, which does not constrain the coefficients of the independent variables, to multiple outputs.
	
	Next we present the solution to~(\ref{eqn:cls1}).
	First let $v$ be fixed. The problem becomes ordinary least squares in $u$, yielding
	\[
	u=(X^\T X)^{-1}X^\T Yv.
	\]
	Let $H=I-X(X^\T X)^{-1} X^\T$, a symmetric idempotent matrix.
	After substituting for $u$, the problem in $v$ is given by
	\[
	{\underset{v\in\R^{d_Y}}{\min}} \|HYv\|_2^2 \quad \text{\quad subject to\quad}  v^\T v = 1.
	\]
	This problem resembles PCA except with a minimum instead of maximum. The solution $v$ is the eigenvector with the lowest eigenvalue of $Y^\T H^\T H Y = Y^\T H Y$.

	%

	\subsection{Multiple Components}
	In CCA, subsequent canonical variables are uncorrelated with each other. After changing these constraints  to be independent of the data, we are left with simple orthogonality constraints between vectors of coefficients. The generalization of~(\ref{eqn:cls1}) to $m$ components is then
	
	\begin{equation}
	\label{eqn:cls2} 
	\begin{aligned}
	& \min\limits_{\substack{U\in\R^{d_X\times m} \\ V\in\R^{d_Y\times m}}}
	& & \|XU-YV\|_\F^2 \\
	& \text{\quad subject to}
	& &  V^{\T} V = I.
	\end{aligned}
	\end{equation}
	
	Again, this problem has a positive semidefinite objective but quadratic constraints. It is difficult to solve analytically because all components must be found simultaneously. 
	We instead choose an easier suboptimal solution: let $V$ be the eigenvectors corresponding to the $m$ lowest eigenvalues from the solution to~(\ref{eqn:cls1}), and compute $U$ accordingly.
	This solution corresponds to greedily solving for each component sequentially under orthogonality.
	The computational runtime of this algorithm is $O(n(d_X + d_Y)^2)$.
	It is an interesting tangent to juxtapose this procedure with Principal Components Analysis (PCA), which solves a similar problem
	\begin{equation*}
	\underset{W \in \R^{d\times d}}\max \|ZW\|^2_\F
	\text{\quad subject to\quad}  W^{\T} W = I
	\end{equation*}
	where $Z \in \R^{n\times d}$ is a centered data matrix.
	In PCA, the greedy eigenvector solution is optimal because of the orthogonality constraints between full vectors of coefficients.
	In CLS, however, only the vectors $v_i$ must be orthogonal, rendering the greedy solution suboptimal.
	
	Separately, in the special case that $m=\min\lbrace d_X,d_Y\rbrace$, then $U$ or $V$ is an orthogonal matrix, so CLS reduces to ordinary least squares on the columns of $X$ or $Y$ respectively.
	
	%
	%

	\subsection{Clustering}
	So far we have presented how to change CCA, a multiple correlation problem, to CLS, which can be thought of as a multi-output regression. On its own, CLS is probably uninteresting, but it becomes relevant in the context of clustering. We assume the observations in a dataset have a nonlinear relationship between views, which we approximate by a mixture of linear relationships. The goal is to compute the cluster assignments and corresponding linear relationships simultaneously. We do so by assigning observations to different CLS structures, which are all simultaneously learned.
	Our proposed CLS clustering algorithm takes data matrices $X$ and $Y$, a number $k$ of clusters, and a number $m$ of components. Let $X^{(i)}$ and $Y^{(i)}$ denote $X$ and $Y$ with rows sub-sampled to those in cluster $i$. Let the coefficients corresponding to that cluster be $U^{(i)}$ and $V^{(i)}$. To find cluster labels for each data point, we iterate the following steps until convergence:
	\begin{itemize}
		\item \textbf{CLS step}\quad Given cluster labels, for each cluster $i=1,\dots,k$: run CLS (\ref{eqn:cls2}) on $X^{(i)}$ and $Y^{(i)}$ to find $U^{(i)}$ and $V^{(i)}$.
		\item \textbf{Labeling step}\quad Given CLS coefficients $U^{(i)}$ and $V^{i)}$,
		for each observation $(x_\ell,y_\ell)$, $\ell=1,\dots,n$: assign it to 
		\[
		\argmin_i \|y_\ell^\T V^{(i)}  - x_\ell^\T U^{(i)}\|_2^2.
		\]
	\end{itemize}
	This procedure takes a block coordinate-wise iterative approach,  resembling Expectation-Maximization~\cite{Dempster1977}, to solving the overall optimization problem
	\begin{equation}  \label{eqn:cls3}
	\begin{aligned}
	& \sum_i  \underset{V^{(i)}\in\R^{d_2 \times m}}{\underset{U^{(i)}\in\R^{d_1 \times m}}{\min}} \|R^{(i)}(XU^{(i)}-YV^{(i)})\|_\mathcal{F}^2\\
	& \text{\quad subject to\quad}  V^{(i)\T} V^{(i)} = I, \; i=1,\dots,k,
	\end{aligned}
	\end{equation}
	where $R^{(i)}$ is a length $n$ diagonal matrix whose $\ell$-th diagonal element is the binary indicator of whether observation $\ell$ is assigned to cluster $i$.
	
	Convergence is guaranteed when $m=1$, i.e., when only the first pair of components is used. 
	The CLS step optimizes over the $u_i$'s and $v_i$'s, while the labeling step optimizes over the $R^{(i)}$'s. 
	Thus the objective is non-increasing at every step, so convergence is guaranteed.
	If $m>1$, an exact solution to CLS would also guarantee monotonicity, but since a greedy approximation  is used, monotonicity is not guaranteed. Nevertheless, we have found the objective function to almost always behave monotonic empirically.

	
	A similar clustering algorithm was proposed by~\cite{Fern2005} called CCA clustering, which considers observations to be generated by a mixture of CCA generative models~\cite{Bach2006}. The algorithm also iterates between an assignment step and CCA step.
	However, while CCA maximizes correlation between variables in the latent space, CLS minimizes the squared error. 
	These objectives are similar, but CLS can find components with weaker correlation and smaller residuals, which is not necessarily an advantage or disadvantage.
	Moreover, CLS clustering solves one important issue with CCA clustering. Intuitively, in CCA clustering the overall clustering objective differs from the CCA objective.
	Recall that the CCA optimization had constraints dependent on data,
	\[
	u^\top X^\top X u = v^\top Y^\top Y v = 1.
	\]
	As a result, when cluster assignments change, the constraints for each cluster's CCA problem change as well.
	To be consistent, the search space for cluster assignments would also have to satisfy those constraints, but this requirement is infeasible.
	Consequently, there is no reason for the CCA clustering algorithm to improve its solution at each iteration.
	By removing the dependence on data in constraints, CLS clustering avoids this problem and therefore permits a more well-behaved optimization routine.
	Indeed, we conducted simple simulations on synthetic Gaussian data and found that CCA clustering often finds poor solutions whereas CLS clustering finds reasonable solutions far more often.

	\subsection{Classification}
	\label{sec:classification}
	Our clustering method can be considered a nonlinear way to characterize multi-view relationships. An unsupervised method, it largely serves descriptive analytics purposes. Yet it is natural to hypothesize that these multi-view relationships can serve as discriminative factors in supervised learning. To the best of our knowledge, this idea is novel.
	Here we propose a straightforward way to build a supervised classification method on top of CLS clusters.
	First CLS clusters are learned independently on each class.
	Then new points are scored for each class according to the best fitting (lowest scoring) cluster in that class's fitted model.
	More formally, the score, or loss, of point $(x,y)$ in cluster $i$ is given by
	\[
	\frac 1 2 \| x^\top U^{(i)} - y^\top V^{(i)} \|_\F^2.
	\]
	The minimum of these scores is taken over the clusters in a given class to produce the score for that class.
	The final classification is the class that has the best fitting cluster overall.
	This approach is intuitively similar to the nearest-neighbor algorithm but in a multi-view relationship space.
	In addition, the procedure can be run many times with different random initializations and the scores averaged, which would make this classifier an ensemble method.
	Although this method differs from most classifiers because it does not maximize separation between classes, it can be understood as a multiclass anomaly detection method, since anomaly detection generally does not maximize separation either.
	
	We next present a way to interpret this classifier's decisions in terms of individual features when there are only two classes. In many applications, it is interesting to examine only two of the learned clusters and ask how to decide which of them a new observation should belong to. The idea is to derive a locally linear model of the relevant factors, which should be readily interpretable. It must be noted, however, that the linearization is imprecise because it loses crucial multi-view information, so this approach should be used with caution.  
	Let the observation be given by $(x,y)$ and let $z$ be the vertical concatenation of $x$ and $y$. 
	Let the two clusters of interest have coefficients $U^{(i)}$ and $V^{(i)}$, where $i \in \{0, 1\}$, and let $W^{(i)}$ be the vertical concatenation of $U^{(i)}$ and $-V^{(i)}$. 
	The loss in cluster $i$ is then $z^\T W^{(i)} W^{(i)\T} z/2$.
	The classification score between the two clusters is  \[\frac 1 2 z^\T (W^{(0)} W^{(0)^\T} - W^{(1)} W^{(1)^\T}) z\] where a higher score indicates membership in cluster 1.
	We determine the effect of a small change in any individual feature by computing the gradient,
	\[
	(W^{(0)} W^{(0)^\T} - W^{(1)} W^{(1)^\T}) z.
	\]
	Thus we can hypothesize about how an observation's features might need to change to alter its classification.

	\subsection{Practicalities}
	\noindent\paragraph{Intercept} An intercept should be incorporated in CLS clustering by augmenting $X$ with a column of $1$'s. 
	
	\noindent\paragraph{Data scale} CCA is affine-invariant with respect to $X$ and $Y$. However, CLS is sensitive to scaling because it uses Euclidean distance, similar to $k$-means. Therefore, we recommend normalizing the column variance in preprocessing.
	
	\noindent\paragraph{Initialization}
	Like in all greedy iterative algorithms similar to $k$-means, random initialization over many runs improves the chance of CLS clustering to reach a robust solution. 
	
	\noindent\paragraph{Appropriate data type} 
	In elided experiments we tested CLS clustering on many different types of synthetic data, which we summarize here. The most appropriate type was found to be continuous values with no missing data. Some distributions on which it performed included Gaussian, log-normal, and uniform. On missing or sparse data, the method requires additional treatment such as imputation before or during the learning process.
	
	\section{Results}
	
	\label{sec:results}
	\subsection{Synthetic Dataset}
	The clustering method was briefly tested on synthetic data to explore its quantitative performance under simple conditions.
	\subsubsection{Description}
	
	We deployed different clustering methods on a medium-sized synthetic dataset. The dataset consisted of 10 equally sized clusters of 1,000 points each. Each cluster was sampled from a different multivariate Gaussian in $\R^{100}$ centered at the origin with covariance drawn from a Wishart distribution. The first 50 features composed one view while rest composed the other view.

	\subsubsection{Results}
	
	\begin{table}[ht!]
		\caption{Cluster Quality on Synthetic Data}
		\label{tbl:cls-synthetic}
		\begin{center}
			\begin{tabular}{lrrrr}
				& CLS    & CCA  & $k$-means     & SC           \\
				\hline \\
				ARI      & .99 $\pm$ .01 &.94 $\pm$ .02 & .005 $\pm$ .003&.000$\pm$ .000
			\end{tabular}
		\end{center}
	\end{table}
	
	We computed the Adjusted Rand Index (ARI) relative to the true cluster labels~\cite{Yeung2001}. We compared CLS clustering to  CCA clustering, $k$-means, and Gaussian kernel spectral clustering (SC). The simulation was run 50 times. Table~\ref{tbl:cls-synthetic} displays the average ARI and two times the standard error. The best performer was  CLS clustering. The next best was CCA clustering, which usually produced solutions with ARI of either about 1.00 or .89 possibly because of convergence issues. Lastly, $k$-means and SC performed very poorly because they searched for isolated $L_2$ clusters structure while the true clusters overlapped.

	\subsection{Medical Dataset}

	\begin{figure}[h]
		\centering
		\includegraphics[width=.35\textwidth]{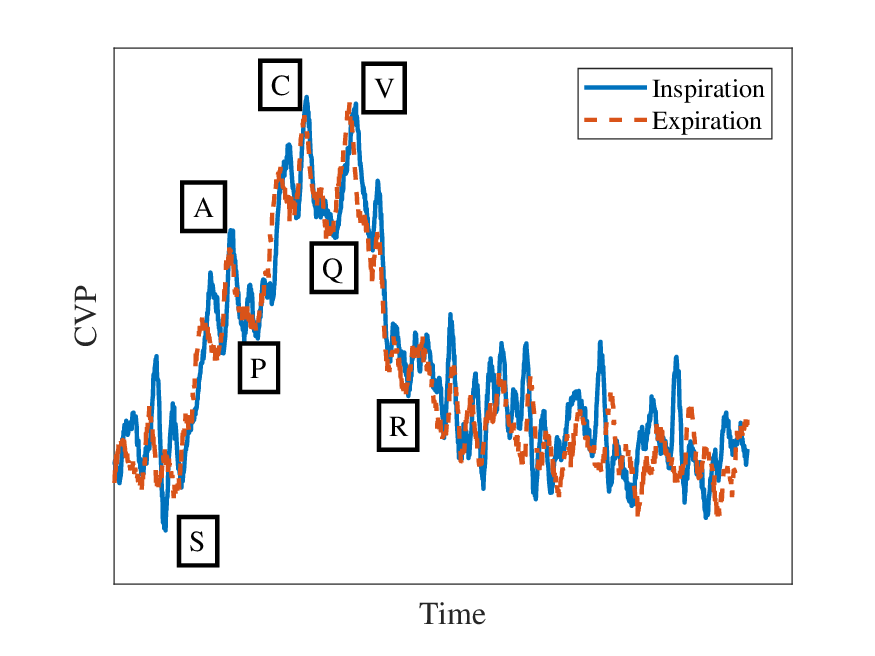}
		\caption{An example of a central venous pressure waveform for inspiration and expiration phases of breathing cycle, along with labeled key points.}
		\label{fig:cvp-waveform}
	\end{figure}
	
	For a qualitative evaluation of CLS clustering, we used a medical dataset to conduct three sets of experiments corresponding to different levels of supervision.
	A primary difficulty in correlation clustering research is that to our knowledge, there is no consensus on a framework for quantitative evaluation. Ideally, our method's cluster quality would be numerically compared on real datasets to alternatives such as $k$-means, spectral clustering, correlational spectral clustering~\cite{Blaschko2008}, CCA clustering~\cite{Fern2005}, and copula-based dependency-seeking clustering~\cite{Rey2012}. However, there are a couple underlying issues with such comparisons. First, the former three alternatives are based on spatial relationships, which makes them improper comparisons because they identify fundamentally different cluster variables. The second issue is the shortage of ground truth in public datasets for correlation clusters. Consequently, in correlation clustering literature, it is common practice to perform qualitative evaluation of clusters rather than quantitative~\cite{Fern2005,Rey2012,Zimek2009}. We do the same in this work.
	\subsubsection{Background}

	We considered a dataset in which we attempted to detect the presence of bleeding and other conditions by monitoring central venous pressure (CVP), the blood pressure measured invasively in the central veins close to the heart, or estimated from indirect less invasive measurements. It helps quantify right atrial pressure and can be used as an estimate of right ventricular preload, important diagnostic tools.
	Predictive tasks based on CVP have been the subject of several studies in the medical literature~\cite{michard2000using,pinsky2005functional,kumar2004pulmonary,marik2013does,damman2009increased,boyd2011fluid}.
	Here we investigate the CVP signal within a controlled setting by attempting to classify CVP waveforms as indicative of an active bleeding episode vs.\ periods of no-bleeding.
	We show how CLS can make predictions as well as automate the discovery of insights of potential clinical interest.
	CLS clusters can be interpreted as clinical phenotypes characterizing patients' pre-bleeding or post-bleeding responses.
	Also, the relationship of bleeding with inspiration and expiration can be interpreted in terms of the original CVP waveforms.
	
	\subsubsection{Description}
	
	The data were collected from an experiment in which healthy  pigs were subjected to controlled bleeding.
	The experimental procedure was similar to that in~\cite{pinsky1984instantaneous}. Thirty-eight Yorkshire pigs were anesthetized, instrumented with catheters, and allowed to stabilize for 30 minutes. Then they were bled at a constant rate of 20 mL/min. Their CVP was monitored for 25 minutes before bleeding and 25 minutes after its onset. Two CVP waveforms (Fig.~\ref{fig:cvp-waveform}) were extracted from each respiration cycle, one from the top of the inspiration phase of breathing and the other from the bottom of expiration. The respiration cycles lasted 5.2 seconds each on average, resulting in an average of 556 observations per pig over the 50 minutes of observation. Thirteen features were extracted from each waveform as averages and ratios between different characteristic points of the CVP waveform, landmarks used commonly in clinical analysis. These features included differences between peaks and troughs such as the height $SA$ between points $S$ and $A$ as well as ratios of ranges such as $VR$ over $CQ$ (Fig.~\ref{fig:cvp-waveform}).

	\begin{figure}[t!]
		\centering
		\includegraphics[width=.38\textwidth]{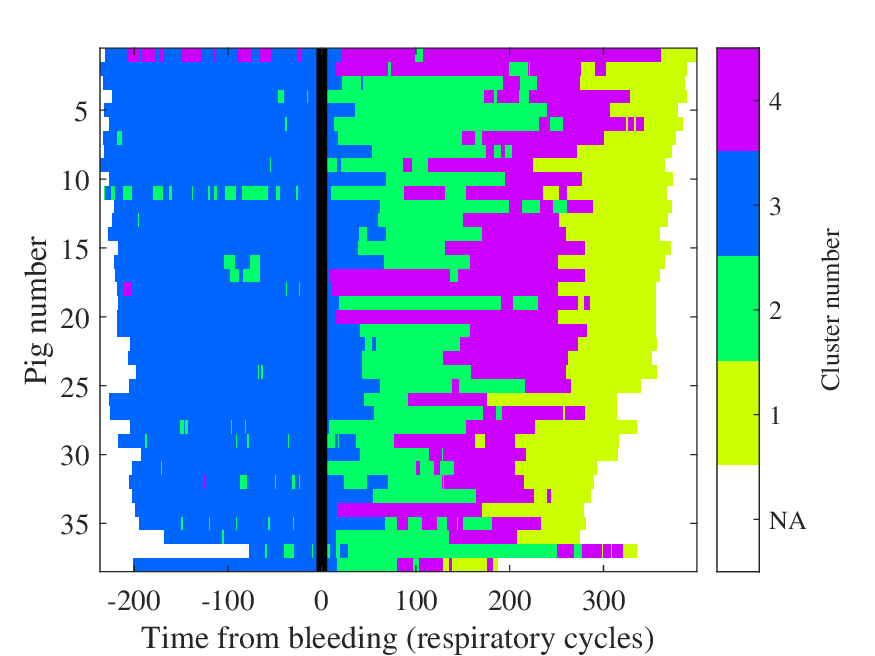}
		\caption{CLS cluster assignments with 4 clusters when blood loss is known.}
		\label{fig:cls-sup-k4}

		\includegraphics[width=.38\textwidth]{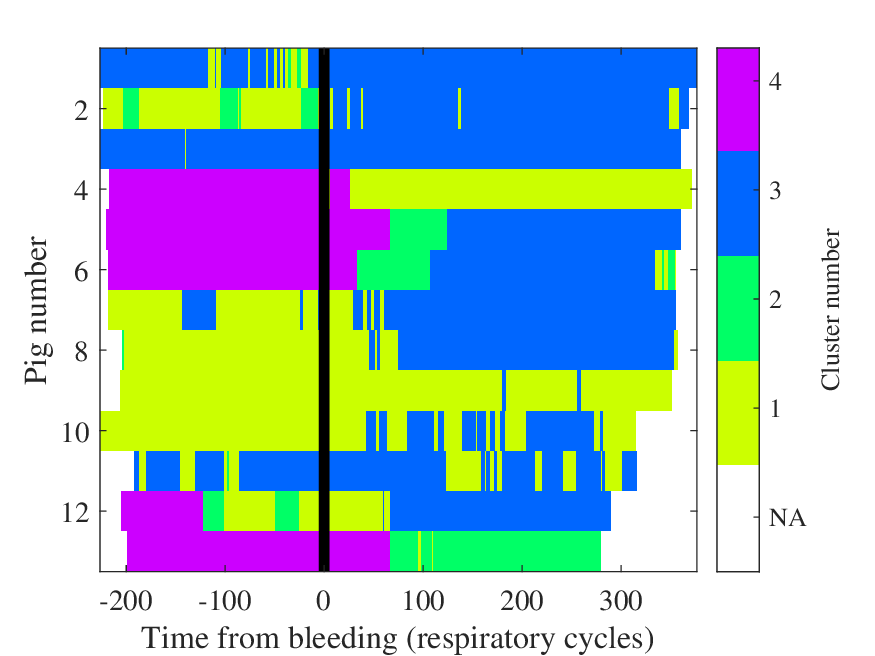}
		\caption{Cluster assignments from $k$-means with 4 clusters when blood loss is unknown.}
		\label{fig:pig-kmeans-unsup-k4}
		
		\includegraphics[width=.38\textwidth]{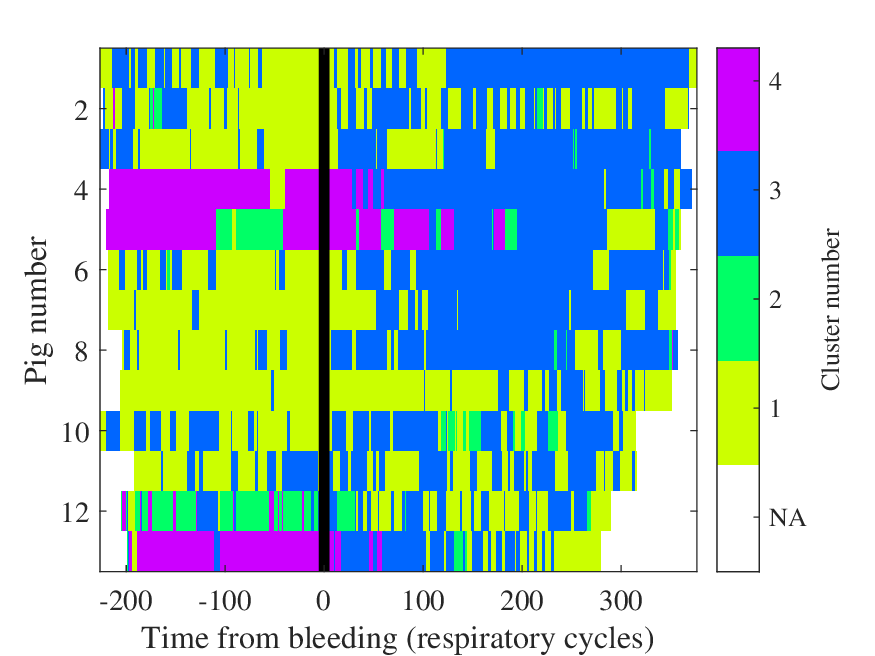}
		\caption{CLS cluster assignments with 4 clusters when blood loss is unknown.}
		\label{fig:cls-clusters-unsup-k4}
	\end{figure}
	

	\subsubsection{Blood Loss Known Exactly} 
	In the first experiment, the exact amount of blood loss was assumed to be known as zero before bleeding and linear at a rate of 20 mL/min after the onset of induced bleeding. CLS clustering was run to cluster respiration cycles with all CVP features as the input view and blood loss as the output view. The data from all 38 pigs were concatenated. Since the output view was univariate, the method corresponded to cluster-wise linear regression~\cite{spath1982fast}.
	The purpose of this experiment was to learn clusters that corresponded to bleeding status by directly incorporating bleeding information. 
	We tried $k=4$ and $m=4$. These values were selected by hand to optimize visual quality of the clusters.
	The cluster assignments are shown in Fig.~\ref{fig:cls-sup-k4}. Each row represents a subject, while each column represents a time step. The clusters are color-coded. 
	The clusters were largely contiguous in time, even though there was no such constraint in the method.
	One cluster corresponded to no bleeding, but the bleeding period was separated into three separate phases. 
	This result confirms the hypothesis that the chosen parameterization of the CVP waveform carries in its structure the information about the bleeding status of the subject and is to some extent informative of the amount of blood lost.
	
	\begin{figure}[t!]
		\centering
		\includegraphics[width=.38\textwidth]{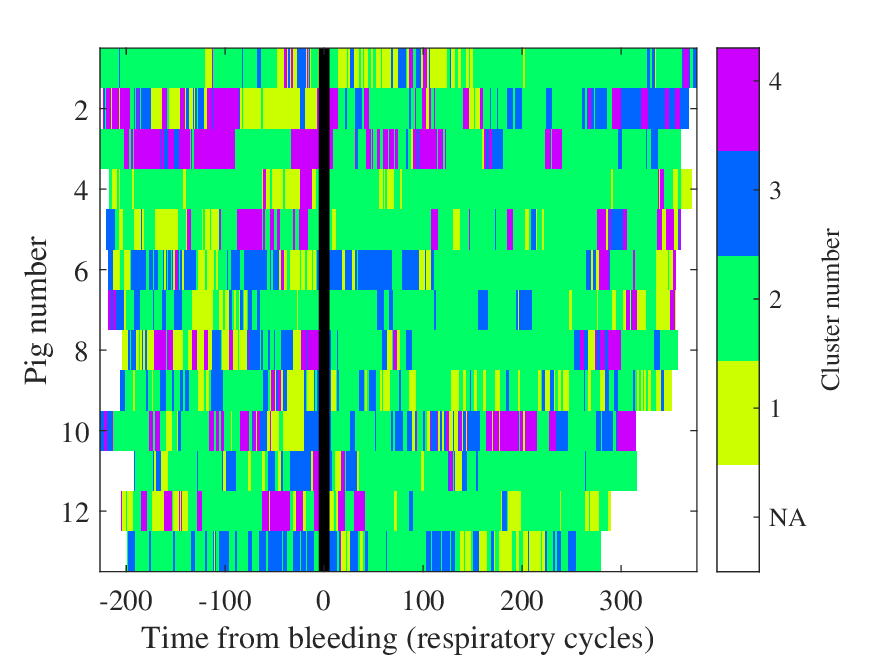}
		\caption{CCA cluster assignments with 4 clusters when blood loss is unknown.}
		\label{fig:cca-pig-unsup-k4}.
		
		\includegraphics[width=.5\textwidth]{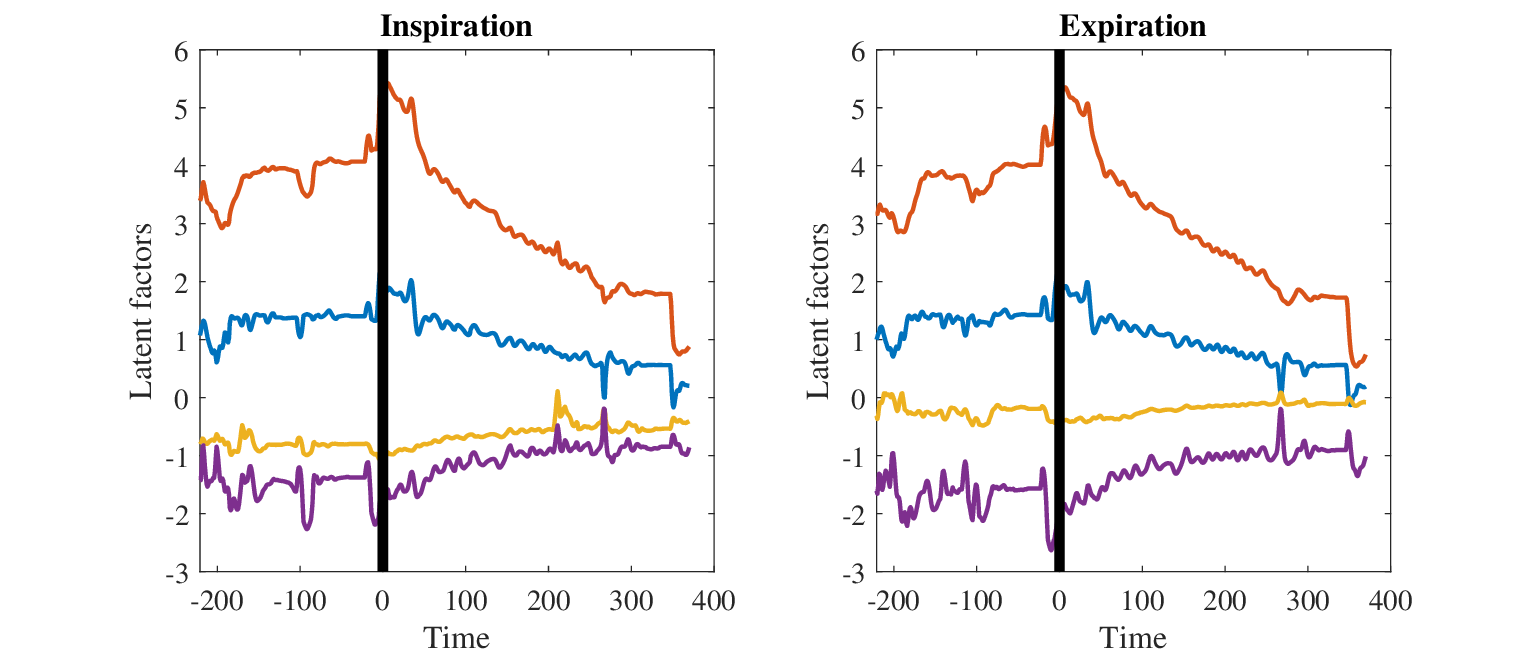}
		\caption{CLS latent components over time for one cluster when blood loss is unknown.}
		\label{fig:cls-unsup-pig7-factor-timeseries-cluster4}
		
		\includegraphics[width=.38\textwidth]{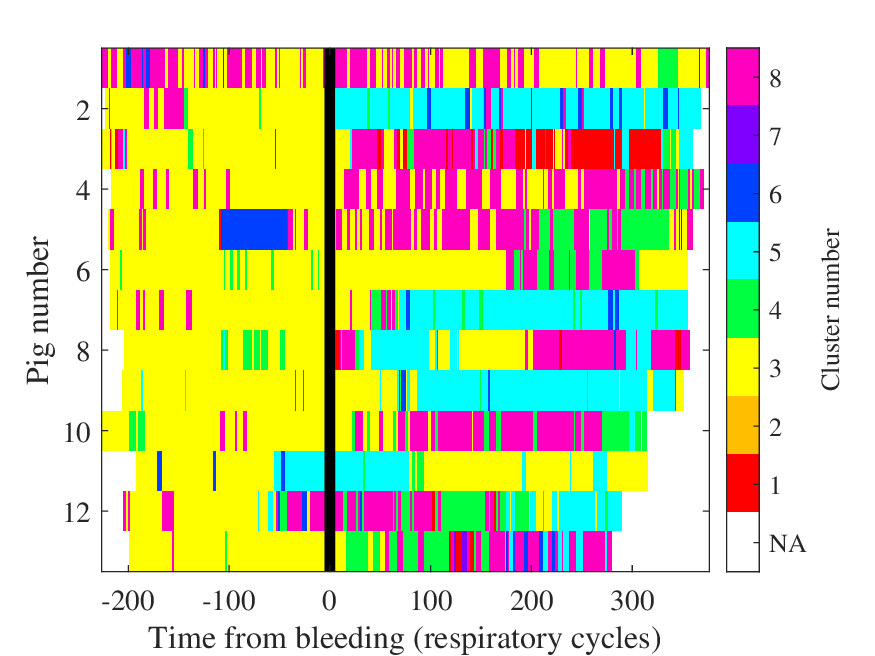}
		\caption{CLS cluster assignments with 3 non-bleeding (1-3) and 5 bleeding (4-8) clusters when blood loss is known as a binary label.}
		\label{fig:cls-ensemble-test-pig-labels}
	\end{figure}

	\begin{figure*}[t]
		\centering
		\includegraphics[width=.8\textwidth]{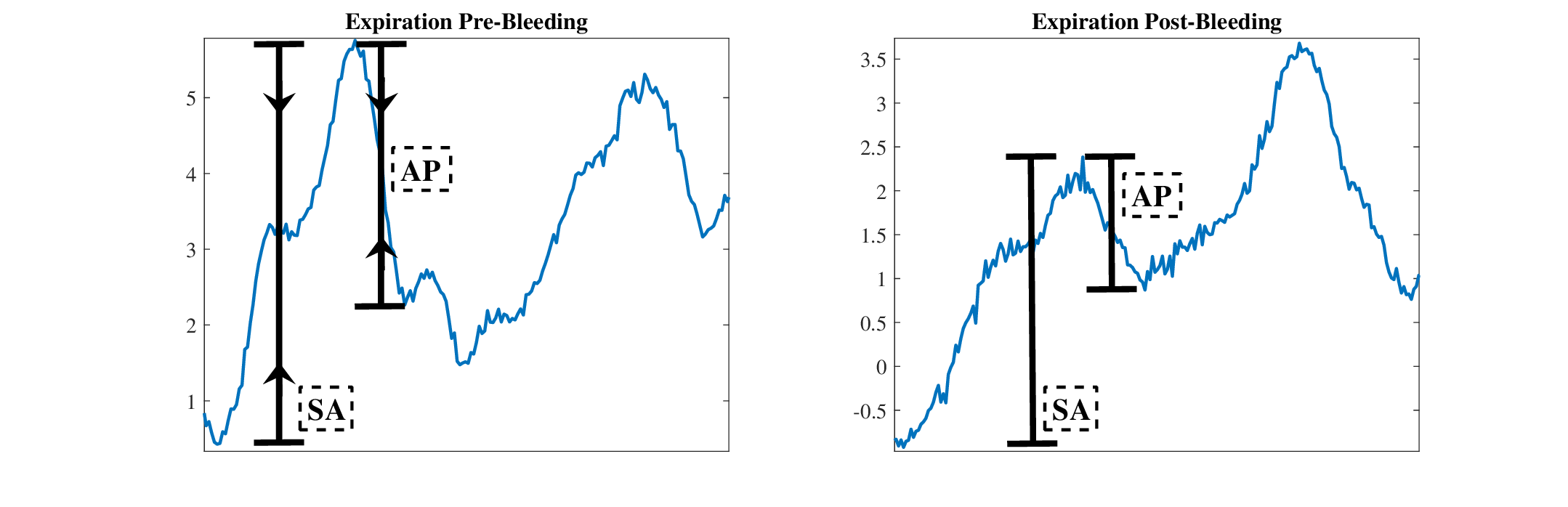}
		\caption{A pair of CVP waveforms from expiration before and after bleeding. The impact of certain features has been labeled on the pre-bleeding side. Arrows indicate lengths that must decrease to appear more like a waveform from after bleeding.}
		\label{fig:cls-pig3-cluster1-comp6-grad}
	\end{figure*}

	\subsubsection{Blood Loss Unknown} 
	In practice, the amount of blood loss is unknown, but this information is required at test time to cluster new observations. To simulate a more practical environment, the second experiment was to deploy CLS clustering on only CVP features without knowing blood loss. The two views were inspiration and expiration, both known at test time. The rationale for these views was that breathing directly alters CVP by varying intrathoracic pressure, transiently changing the pressure gradient for blood flow back to the heart.  Thus, we reason that features of CVP during inspiration and expiration should be informative in early identification of emerging changes of bleeding status.
	Moreover, by defining views as temporal windows about regular events, the relationship between them can be analyzed as a signal of temporal change in physiological state.
	
	Observations from the same pig were constrained to belong to the same cluster during training.
	The pigs were partitioned into training and test sets of 25 and 13 subjects respectively. 
	Data from all training pigs were concatenated to learn the cluster model.
	The purpose of this experiment was to examine the qualitative performance of this method in more realistic conditions. 
	We chose $k=4$ and $m=4$ to match the previous experiment.
	
	The cluster assignments are shown in Fig.~\ref{fig:cls-clusters-unsup-k4}. The clusters were still mostly time-contiguous, although they appeared significantly noisier than before. This result was expected because critical information, the amount of bleeding, was excluded. Two clusters corresponded predominantly to either no bleeding or bleeding, but the two other clusters identified two small groups of pigs that appeared distinctive before the onset of bleeding. This suggests that our method discovered diversity within the subjects' physiology in the stable state, later confirmed via independent review.
	Crucially, the clusters still corresponded to bleeding status even though no information about bleeding was known.
	For comparison, Fig.~\ref{fig:pig-kmeans-unsup-k4} displays the clusters from $k$-means on the concatenated views. The clusters are time contiguous and smoother than CLS clusters, which is expected because observations close in time are probably close in $L_2$ distance. Overall, the two approaches find similar cluster structures. This result suggests that bleeding status is expressed in both correlation between views and $L_2$ distance. However, for some pigs the CLS clusters have a clearer divide between no bleeding and bleeding, such as pigs 1, 3, and 11. In addition, pig 4 is notable because $k$-means does not differentiate between its bleeding and no-bleeding states, whereas for CLS clustering that distinction is more clear.
	It is plausible that $k$-means is underfitting the clusters, while the noise in the CLS clusters indicates that our method may be further from underfitting (and closer to overfitting).
	
	Additionally, Fig.~\ref{fig:cca-pig-unsup-k4} shows for comparison the clusters from CCA clustering on inspiration and expiration views. The same constraints and number of clusters and components were used as in CLS clustering, yet the clusters do not appear much time-contiguous and do not correspond  with bleeding status.
	Also, we clustered the same data using spectral clustering~\cite{VonLuxburg2007} using a Gaussian kernel to produce the affinity matrix, but over 98\% of data points were assigned to the same cluster.
	
	Fig.~\ref{fig:cls-unsup-pig7-factor-timeseries-cluster4} shows the four latent components over time of one pig in a particular cluster. The components appear almost constant before bleeding and become more or less linear after bleeding. This behavior is interesting because it resembles the amount of blood loss over time, even though this information was excluded from training. Although this example is from one cluster, it is also representative of all other clusters.
	
	\subsubsection{Blood Loss Known as Binary}
	
	\begin{table}[t]
		\caption{Bleeding Classification Performance}
		\label{tbl:perf}
		\begin{center}
			\begin{tabular}{rccc}
				\multicolumn{1}{r}{}& \multicolumn{1}{c}{Single cluster CLS}  &\multicolumn{1}{c}{Final CLS} &\multicolumn{1}{c}{Random forest}\\
				\hline \\
				AUC          & .701 $\pm$ .128      & {.862} $\pm$ .064 & {.891} $\pm$ .075\\
				TPR @ .10 FPR & .468 $\pm$ .185      & {.674} $\pm$ .145 & {.762} $\pm$ .167\\
				TPR @ .01 FPR & .222 $\pm$ .134      & .501 $\pm$ .185 & {.610} $\pm$ .210\\
				FPR @ .50 TPR & .239 $\pm$ .152      & {.064} $\pm$ .055 & .073 $\pm$ .075\\
			\end{tabular}
		\end{center}
	\end{table}
	
	
	In the unsupervised setting, it was difficult to obtain quantitative measures because there was no ground truth.
	Thus, the third experiment was to run CLS classification to determine whether the learned multi-view structure had any discriminative power. The classification task was to decide whether an observation came from before or after the onset of bleeding. The binary label is required at training time but is not needed to classify or cluster unseen observations, so this form of supervision is more practical than the first.
	Under the same training/test split, data from all training pigs were concatenated to learn the cluster model.
	Leave-one-subject-out cross-validation was employed to select the hyperparameters $k$ and $m$ for non-bleeding and bleeding models.
	Following the classification algorithm from Sec.~\ref{sec:classification}, CLS clusters were learned separately on the two classes.
	Observations from the same pig were constrained to belong to the same cluster during training.
	The classification scores on a left-out pig were used to compute the area under the receiver operating characteristic curve (AUC), true positive rate (TPR) at a false positive rate (FPR) of 10\% and 1\%, and FPR at a TPR of 50\%.
	Hyperparameters were selected by optimizing the AUC.
	We chose 3 clusters with 6 components each for pre-bleeding and 5 clusters with 7 components each for post-bleeding.
	
	Fig.~\ref{fig:cls-ensemble-test-pig-labels} illustrates the resulting cluster assignments. Similarly to the previous experiment, the clusters are mostly time contiguous but contain substantial noise. There is one predominant cluster for no bleeding and several for bleeding. Intriguingly, the cluster structure does not appear too similar to the previous experiment and has more differentiation between subjects during the bleeding period, suggesting some extent of individualization of the response to harmful effects of bleeding across subjects.
	
	Table~\ref{tbl:perf} shows performance metrics of the final model on test data. 
	It also gives results from a model that learns only one cluster on each class. 
	The sizable gap in performance between single cluster and multiple clusters demonstrates the benefit of searching for correlations that exist in subsets of the data, as opposite to a global correlation model identifiable in the whole set.
	The table includes results from a random forest classifier~\cite{Breiman2001} with 100 trees trained on the combined views.
	The random forest performs best in most metrics, but its advantage vs.\  CLS is not statistically significant.
	This result is acceptable since CLS enables a detailed yet interpretable view of discovered structures in data while its performance metrics remain within the confidence interval of otherwise powerful random forest classifier.
	Our goal was to illustrate that CLS classification was pragmatically close to state-of-the-art of methods used in clinical settings, even if just slightly worse, because it demonstrates that multi-view relationships can effectively serve to discriminate between statuses of bleeding or not bleeding. Furthermore, the advantage of the proposed method is not intended to be classification performance but rather its cluster structure.
	
	To understand the model's decisions, we used the method involving the gradient derived in Sec.~\ref{sec:classification} on weak learners from the pre- and post-bleeding ensembles.
	We checked the score that determined whether a certain pig belonged to cluster 1 or 4, where cluster 1 was pre-bleeding and cluster 4 was post-bleeding.
	We computed the gradient of the score on a pre-bleeding observation.
	The results are displayed in Fig.~\ref{fig:cls-pig3-cluster1-comp6-grad}. 
	The original waveform of the observation is plotted on the left.
	According to the gradient, the most major changes that would make the observation closer to a bleeding waveform were shortening the lengths $SA$ and $AP$ during expiration. 
	Correspondingly, the figure shows on the right an expiration waveform from soon after the onset of bleeding.
	The two characteristic waveform parameters have shrunk dramatically, and bumps and dips at $A$ and $S$ have substantially diminished.

	\section{Discussion}
	\label{sec:discussion}
	
	\subsection{Bleeding Clusters} Figs.~\ref{fig:cls-sup-k4},~\ref{fig:cls-clusters-unsup-k4}, and~\ref{fig:cls-ensemble-test-pig-labels} highlight an interesting pattern. For many pigs, there was a dominant cluster before bleeding, but when bleeding started, a different cluster took over. This new cluster typically only held observations from the first ten or fewer minutes after bleeding. Afterward, other clusters became dominant.
	One interpretation is that the physiological response to bleeding changed as the induced stress escalated. There may have been an initial compensation surge, followed by a more systemic response mediated through autonomic nervous control which could also change in its modality as a function of escalating stress.
	This hypothesis may be supported by Fig.~\ref{fig:cls-unsup-pig7-factor-timeseries-cluster4}, which shows that the immediate onset of bleeding corresponded to a spike in latent variables.
	This pattern is an example of how the interpretable structure of CLS clustering can lend itself to finding practical insights.
	In addition, our method identified diversity among subjects. A few pigs did not follow the majority in their clustering patterns before and during bleed (Fig.\~ref{fig:pig-kmeans-unsup-k4}). 
	
	\subsection{Clinical relevance}
	Many physiologic factors interact to define a given CVP or its mean change during the ventilatory cycle making these measures insensitive to changes in effective circulating blood volume as bleeding occurs except at the extremes, where such monitoring is not needed. Importantly, as depicted in Fig.~\ref{fig:cls-pig3-cluster1-comp6-grad}, the dynamical waveform changes at end-expiration in the CVP waveform features compared to end-inspiration are very informative of dynamic changes in volume status, even if the absolute CVP values are not. This is relevant to bedside monitoring of critically ill patients for several reasons. First, CVP monitoring is common in critically ill patients because central venous catheters safely deliver fluid and drugs that cannot safely be infused by a peripheral source. Thus, its monitoring is readily available. Second, although absolute CVP values may be inaccurate for technical reasons of zero reference values, the pressure waveform datasets remain accurate, allowing their featurization for CLS and other machine learning applications that until now have been underutilized.  And finally, early identification of occult bleeding would allow earlier corrective therapies to minimize or prevent hypovolemia associated tissue hypoperfusion. Such earlier interventions would markedly reduce hypoperfusion related morbidities, like acute kidney injury, ileus and secondary wound infection.

	
	\subsection{Generalization}
	This work focused on a narrow problem in the medical domain, but there exist many opportunities to apply the same methodology in other areas too. 
	As a clustering method, this work offers a nonlinear alternative to CCA to characterize relationships between two views. The key property of our method is its cluster structure, which enables interpretation of the different groups of observations discovered. This methodology could be used in general two-view settings, such as multi-modal data, to learn clusters that operate on characteristics of data fundamentally different from those learned by standard clustering algorithms.
	Also, as a classifier, this work can be generally applied to any two-view classification problem. After training on labeled data, it scores new observations in each class. This approach can be useful in problems where relationships between two views are hypothesized to indicate differences between classes, which may include a multitude of applications. Indeed, in our experiments outside this work, we observed strong quantitative performance of this method on several real-world datasets.
	In addition, even though it does not explicitly leverage temporal structure, our method's application of multi-view relationships to time series analysis can be transferred to virtually any temporal data to analyze variability temporally and cross-sectionally. To recap the approach, views are represented by windows before and after regular events in order to characterize dynamics of temporal change. Features are computed for each window through various time series featurizations. Then our clustering and classification routines can be applied. This method can be useful to examine time series whose changes over time are governed by a nonstationary process.
	This direction also inspires future work to modify the clustering method to find temporally smooth clusters. A potential solution would be to do the assignment step in temporal order and bias the next point's label toward the previous one.

	\subsection{Soft Clustering} We developed a soft clustering extension of CLS based on ideas in~\cite{Hathaway1993}.
	One way to view this extension is that cluster probabilities of an observation are regularized toward a uniform distribution over clusters.
	In the soft version, the optimization is much smoother, resulting in more consistent solutions over different runs. In the applications shown in this paper, however, it was outperformed by the hard version, even though the assignment step in the hard version is highly non-smooth. A potential avenue for future work would be to analyze this and other theoretical optimization properties of the method.
	

	\section{Conclusion}
	
	\label{sec:conclusion}
	This work considered the problem of discovering explainable structures in complex multi-view datasets.
	Our main approach was to consider relationships between two views explicitly as a unit of analysis.
	We proposed a method to characterize nonlinear relationships between two views by modeling them as a mixture of linear relationships. This method could be considered a multi-view correlation clustering algorithm.
	We also proposed a routine to perform supervised classification using the discovered correlation clusters as a basis. Intuitively, this classifier exploits differences in multi-view relationships between classes.
	The method was tested on CVP waveform datasets of induced bleeding and was demonstrated to perform well.
	Furthermore, the experiments showcased the method's capability  to discover interesting patterns and produce explanations of its predictions.
	We demonstrated the potential utility of the proposed CLS method on the task of real-time monitoring of surgical patients, but it can be useful in a wide range of multi-view problems in clinical and biological engineering applications,
	wherever distinct structures of relationships between views of data can reveal operationally useful information.
	
	\section*{Acknowledgment}
	This work was partially supported by DARPA under award FA8750-17-2-0130 and by National Institutes of Health under award GM117622.
	%
	
	
	\bibliographystyle{ieeetr}
	\bibliography{refs,CorrelationClustering,QF}
	
	
\end{document}